\documentclass{article}

\usepackage{microtype}
\usepackage{graphicx}
\usepackage{subfigure}
\usepackage{booktabs} % for professional tables

\usepackage{hyperref}

\usepackage[accepted]{arxivlocal2024}
\usepackage{amsmath}
\usepackage{amssymb}
\usepackage{mathtools}
\usepackage{amsthm}
\usepackage{multirow} 

\usepackage[capitalize,noabbrev]{cleveref}

\newcommand{\secref}[1]{Sec. \ref{#1}}
\newcommand{\figref}[1]{Figure \ref{#1}}
\newcommand{\eqnref}[1]{Equation \ref{#1}}
\newcommand{\tabref}[1]{Table \ref{#1}}

\newcommand{\sArt}[0]{state-of-the-art}

\newcommand{\R}[0]{\mathbb{R}}
\newcommand{\Q}[0]{\mathbb{Q}}
\newcommand{\N}[0]{\mathcal{N}}
\newcommand{\B}[0]{\mathcal{B}}
\newcommand{\ie}[0]{i.e.}
\newcommand{\papertitle}[0]{Minimize Quantization Output Error with Bias Compensation}

\theoremstyle{plain}

\theoremstyle{definition}

\theoremstyle{remark}

\usepackage[textsize=tiny]{todonotes}

\icmltitlerunning{\papertitle}

\begin{document}

\twocolumn[
% \icmltitle{Submission and Formatting Instructions for \\
%            International Conference on Machine Learning (ICML 2024)}
\icmltitle{\papertitle}



\icmlsetsymbol{equal}{*}

\begin{icmlauthorlist}
\icmlauthor{Cheng Gong}{yyy}
\icmlauthor{Haoshuai Zheng}{xxx}
\icmlauthor{Mengting Hu}{yyy}
\icmlauthor{Zheng Lin}{xxx}
\icmlauthor{Deng-Ping Fan}{comp,xxx}
\icmlauthor{Yuzhi Zhang}{yyy}
\icmlauthor{Tao Li}{xxx}
\end{icmlauthorlist}

\icmlaffiliation{yyy}{College of Software, Nankai University, Tianjin, China}
\icmlaffiliation{xxx}{College of Computer Science, Nankai University, Tianjin, China}
\icmlaffiliation{comp}{Nankai International Advanced Research Institute (SHENZHEN FUTIAN),Nankai University, Shenzhen, 518045, China}

\icmlcorrespondingauthor{Tao Li}{litao@nankai.edu.cn}

\icmlkeywords{Quantization, Bias Compensation, Output Error, Convex Optimization}

\vskip 0.3in
]



\printAffiliationsAndNotice{}  % leave blank if no need to mention equal contribution
\begin{abstract}
    Quantization is a promising method that reduces memory usage and computational intensity of Deep Neural Networks (DNNs), but it often leads to significant output error that hinder model deployment.
    % Previous studies have made tremendous achievements in minimizing output error. 
    % However, these optimizations are non-convex and pose difficulties in further reducing output error for ultra-low-precision quantization.
    In this paper, we propose Bias Compensation (BC) to minimize the output error, thus realizing ultra-low-precision quantization without model fine-tuning.
    % 不同于以往方法试图去优化非凸的量化过程以降低输出误差，BC通过直接在输出上添加一个偏置向量来直接降低输出误差
    % 因此，BC是与quantizer设计垂直的一项技术，其可以非常容易地在凸优化下求得最优的解析解，
    % 他可以直接降低量化造成的误差从而提升量化模型在超低精度量化下的表现
    % 此外，由于BC仅在输出上添加偏置向量，其几乎不会增加模型推理延迟和内存占用，
    Instead of optimizing the non-convex quantization process as in most previous methods, the proposed BC bypasses the step to directly minimize the quantizing output error by identifying a bias vector for compensation.
    We have established that the minimization of output error through BC is a convex problem and provides an efficient strategy to procure optimal solutions associated with minimal output error, without the need for training or fine-tuning.
    % Moreover, bias compensation operates orthogonal to quantizer optimization and thus can seamlessly integrate with the existing quantizers to achieve low-precision quantization.
    % With regard to computational demands, bias compensation merely involves adding a bias vector to the output of a quantized module, thereby causing virtually no additional computational load.
    % \gc{Experiments results}
    % 广泛的实验表明，BC可以极大地降低量化，尤其是超低精度量化导致得输出误差，从而使得超低精度量化变为可能。
    % 在实验中，我们使用GPTQ和PTQ4VIT作为基准量化方法，成功实现了2bit和4bit的LLM和VIT模型的量化，并相比于基准模型实现了20%-10%的任务性能提升，并且几乎不增加额外推理延时和内存占用。
    We conduct extensive experiments on Vision Transformer models and Large Language Models, and the results show that our method notably reduces quantization output error, thereby permitting ultra-low-precision post-training quantization and enhancing the task performance of models.
    Especially, BC improves the accuracy of ViT-B$^*$ with 4-bit PTQ4ViT by $\textbf{36.89\%}$ on the ImageNet-1k task, and decreases the perplexity of OPT-350M with 3-bit GPTQ by \textbf{5.97} on WikiText2.
    % By combining the proposed bias compensation with the most recent GPTQ and PTQ4VIT, our method attains the 3-bit quantization for and , respectively.
    % Compared with Vallina baselines, our method augments task performance by up to 10\% and 30\% at low-precision weights.
    The code is in https://github.com/GongCheng1919/bias-compensation.
       
\end{abstract}

\section{Introduction}
% \gc{Background-detail}
Quantization is a vital method for efficient neural network computation.
% Quantization is one of the most promising directions as it directly reduces the memory footprint and compute intensity.
It replaces the high-precision floating point weights with low-precision ones, thus saving the memory footprints and computing loads in model inference.
% \gc{Challenges-detail}
The number space of low-precision weights is much smaller than that of float ones, which inevitably results in numerical errors and affects the output of quantized layers.

\begin{figure*}[!ht]
    \centering
    % \vspace{10pt}
    \includegraphics[width=1.0\linewidth]{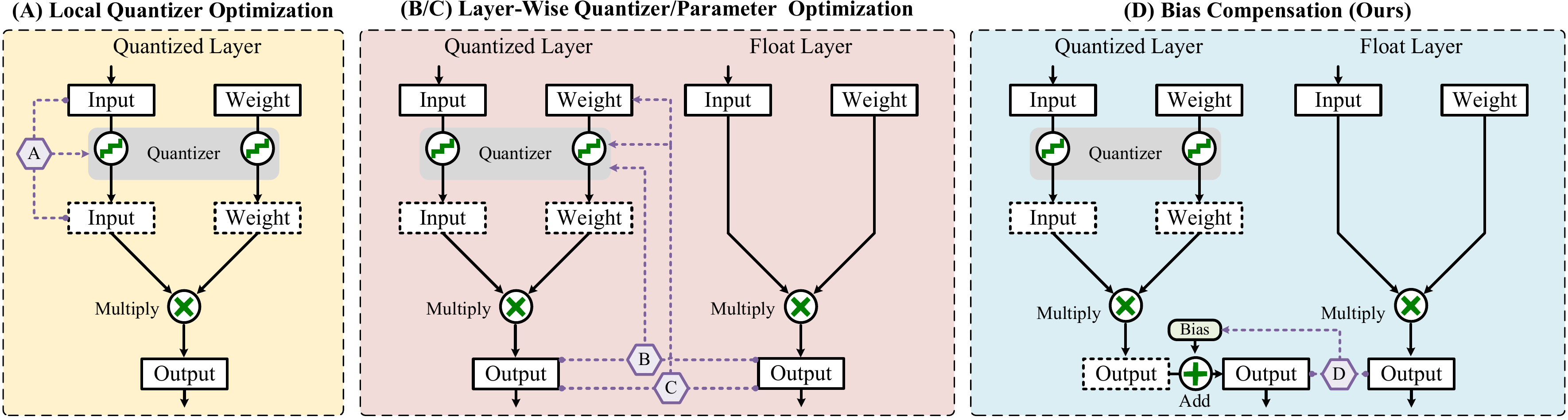}
    \vspace{-15pt}
    \caption{Comparison between the previous PTQ methods and our proposed bias compensation.
    (A) indicates local quantizer optimization methods. (B) and (C) are layer-wise quantizer and parameter optimization methods, respectively.
    Previous methods optimize the quantizer parameter or layer-wise weights to minimize the quantization loss or output error, which is non-convex and difficult to solve. 
    Our method shown in (D) directly minimizes the output error by solving the best bias vector, which is convex and guarantees minimal output error. 
    }
    \vspace{-10pt}
    \label{fig:development-stages-of-PTQ}
\end{figure*}

% \gc{Motivations-detail}
Tremendous achievements have been achieved in minimizing the output errors of quantized layers, and they generally fall into two categories: Quantization Aware Training (QAT) and Post-Training Quantization (PTQ).
% 最相关的工作
% QAT
QAT fine-tunes the quantized DNNs on the whole dataset or trains them from scratch to align the outputs of float models and quantized ones.
It achieves ultra-low-precision (2-3 bits) quantization while maintaining the task performance of models~\cite{gong2020vecq,ESB,liu2023llmqat}, but the full-parameter training on the complete dataset is becoming impractical, especially for large pre-trained models.
% PTQ
% To address this issue, Post-Training Quantization (PTQ) is proposed.
PTQ exhibits great compression efficiency compared to QAT since it is usually applied to quantize the model without retraining and requires no extra dataset or just a small amount of calibration data for adjusting the quantizer and model parameters~\cite{nagel2020adaround,frantar2022gptq,yuan2022ptq4vit}. 
The development of PTQ can be roughly divided into three stages, as shown in \figref{fig:development-stages-of-PTQ}: {local quantizer optimization}, {layer-wise quantizer optimization}, and {layer-wise parameter optimization}.

% local quantizer optimization
Local quantizer optimization reduces the output error by optimizing the local quantization loss offline and finding the local optimal quantizer for each quantized layer in advance~\cite{banner2019post,zhao2019improving,kravchik2019low}. 
% layer-wise quantizer optimization
Layer-wise quantizer optimization adopts a few calibration data to reconstruct the outputs of each layer in models to find a relatively optimal quantizer that can align the outputs of quantized models and float models~\cite{nagel2020adaround,li2020brecq,wang2020bitsplit}.
% layer-wise quantizer and parameter optimization
Layer-wise parameter optimization is similar to layer-wise quantizer optimization, but it updates the quantizer parameters and layer parameters simultaneously to reconstruct the outputs, thus achieving lower precision quantization~\cite{hubara2021adaquant,frantar2022obc,frantar2022gptq}. 
% conclusion
However, these optimizations are usually non-convex,
which makes it difficult to recover the task performance of DNNs from ultra-low precision quantization since large output error.
% ~\cite{frantar2022gptq,yuan2022ptq4vit}.
% and time-consuming. 
% In the third stage, PTQ fine-tunes the full parameters of the quantized layer to recover the task performance of DNNs, and the cost is gradually approaching QAT~\cite{frantar2022obc,yao2022zeroquant}.
% Despite that, compared with QAT, even if full parameter fine-tuning is performed on the calibration dataset, PTQ is still difficult to recover the task performance from ultra-low precision quantization since large output error~\cite{frantar2022gptq,yuan2022ptq4vit}.

% \gc{Our work and Contributions-detail}
In this paper, we minimize the output error caused by quantization from a new perspective of error compensation.
We call it \textit{Bias Compensation (BC)}, which adds a bias vector to the outputs of each quantized layer in models and finds the optimal bias vector for minimizing output error.
We demonstrate that identifying the optimal bias vector for each quantized layer is a convex problem, and solving the optimal solutions for minimal output error is attainable (See \secref{sec:implementation}).
% Additionally, BC is orthogonal to quantizer optimization, and 
It can be easily combined with the existing quantizers and enables ultra-low-precision quantization.
Since the bias vector is added to the outputs after expensive operations, \ie, matrix multiplications,
% or convolutions, 
% the computation results of quantized low-precision tensors, 
it does not affect the computing efficiency of these operations and introduces almost no additional computation loads.
% and does not delay the model inference.
We conduct extensive experiments to verify the validation of BC, and the results show that BC can significantly reduce the output error and improve the task performance of quantized models.

Our contributions are summarized as follows:
% 我们从一个全新的角度出发，提出一种新的优化量化导致的输出误差的方法，称为BC。不同于以往的方法关注于优化非凸的量化过程，我们直接在输出上添加一个可优化的偏置向量，以直接降低输出误差。
% 我们证明对偏置向量的优化是凸的，无需微调即可获得最优的偏置向量，并且证明应用BC可以从理论上保证更低的输出误差。
% 在ViT和LLM上的实验证明，BC显著提升量化模型的任务性能，尤其是保证低精度量化的效果。特别地，
% 
\begin{itemize}
    \setlength{\itemsep}{9pt}
    \item We propose BC for minimizing the quantization-induced output error from a novel perspective. 
    Instead of working on optimizing the non-convex quantization process as in previous methods, BC directly optimizes the output error by adding an optimizable bias vector to the outputs of quantized layers.
    
    \item We prove that the optimization of BC is convex, and the optimal bias vector can be obtained without fine-tuning. Additionally, we theoretically demonstrate that BC can always guarantee lower output error.
    % \item We introduce BC from a novel perspective to reduce the output errors caused by quantization and prove that solving the optimal solutions for the minimal errors is convex and attainable.
    % 我们实现了一个简单的算法来在校准数据集上获取最优偏执补偿矩阵？
    % \item We implement the ultra-low precision quantization of 3 bits by combining BC with \sArt~quantizers.
    \item Experiments on Vision Transformer models and Large Language Models demonstrate that BC can significantly improve the task performance of quantized models.
    % It has enormous potential to enable ultra-low-precision quantization.
    In particular, BC improves the accuracy of ViT-B$^*$ with 4-bit PTQ4ViT by $\textbf{36.89\%}$ on the ImageNet-1k task, and decreases the perplexity of OPT-350M with 3-bit GPTQ by \textbf{5.97} on WikiText2.
    % It consistently improves the perplexity of LLMs with 4-, 3-, and 2-bit weights and the average improvements are $\textbf{3.99}$, $\textbf{364.85}$, and $\textbf{2912.83}$ compared to baselines, respectively.
    % achieve 3-bit quantization for large language models and vision transformer models, respectively, and improve task performance by up to 10\% and 20\% with almost no additional inference delay and memory footprint.
\end{itemize}

% paper structure
\section{Related work}
% 这里有引言暂时没想好写啥，偏差补偿好像是个新的东西，全写量化好像又太偏，还是应该沿着输出误差优化这条路来写
% 沿着输出误差优化这条路写吧
% Quantization replaces high-precision with low-precision representations to save computation loads and memory footprint in DNN inference, while numerical errors are inevitable in replacing.
% The accumulated numerical errors can result in large errors compared to the outputs of corresponding original float models.
% Therefore, minimizing output error is a vital problem in DNN quantization.
% This section presents a brief overview of the achievements in DNN quantization.

\subsection{Quantization Aware Training}
% QAT通过在完整训练集上重新训练来恢复精度。但是QAT这通常消耗巨大，不实用。
QAT fine-tunes the model parameters on the whole dataset to 
recover model accuracy from quantization.
LLM-QAT~\cite{liu2023llmqat} proposes data-free distillation for training quantized models.
QLoRA~\cite{dettmers2023qlora} quantizes weights into 4 bits and employs Low-Rank Adaptation (LoRA)~\cite{hu2021lora}  for fine-tuning.
Block-wise quantization~\cite{dettmers20218bwq} quantizes the states of optimizers for less memory usage in model training.
PEQA~\cite{kim2023memory_peqa} quantizes model weights into 4 bits and fine-tunes only the quantization scales of quantized LLMs. 
Despite tremendous achievements in QAT, fine-tuning the full parameters or even part parameters~\cite{hu2021lora,dettmers2023qlora} on the complete dataset is a heavy workload, and PTQ is studied to address this issue. 
% and the post-training quantization is studied. 
% LLM-QAT~\cite{llmqat} proposes a data-free method to build quantized models.
% It generates data from the float pre-trained model with the next token generation and uses the generated data as input and the float model prediction as the label to guide quantized model fine-tuning.
% QLoRA
% QLoRA~\cite{QLoRA} introduces a new format NormalFloat for weight quantization.
% It quantizes the weights into 4 bits and the quantizer parameters into 8 bits to conserve memory, and employs the low-rank adapter (LoRA) technique to fine-tune quantized models. 
% PEQA
% block-wise quantization 

\subsection{Post-Training Quantization}
% 这里缺乏一些工作，还得去多看看整理一下
PTQ focuses on quantizing the pre-trained models with no data~\cite{banner2019post} or only a small amount of calibration data~\cite{frantar2022gptq}. 
% It does not require learning the full parameters of models on the full dataset, thus saving training time and cost.
According to the type of optimization, the development of PTQ can be roughly divided into three stages as follows:

% \newpage
% \subsubsection{Local Quantizer Optimization}
% 
1) \textbf{Local quantizer optimization} is a greedy strategy that simplifies the minimization of output error to the minimization of quantization loss in \figref{fig:development-stages-of-PTQ}.
% cumulative numerical error in quantization as shown in \figref{fig:development-stages-of-PTQ}.
% It optimizes the quantization loss to indirectly reduce the output error and finally obtain the relatively optimal quantizers for weights and activation.
\cite{banner2019post} employs the analytical clipping range and per-channel mixed-precision quantizer for reducing the quantization loss. 
% It implements the 4-bit quantizers for weight and activation for the mainstream convolutional neural networks. 
% 堆更多工作
OCS\cite{zhao2019improving} reduces the quantization loss by splitting channels into more channels.
% which increases computation but achieves lower precision results in the process.
% 
\cite{kravchik2019low} employs the fine-grainy kernel-wise and multi-tensor quantizer for low loss. 
% Local quantizer optimization is a greedy strategy that indirectly minimizes output error, and it is difficult to further reduce output error.
% but it is not equivalent to it when simultaneously performing activation quantization~\cite{kravchik2019low}. 
% Therefore, local optimization is difficult to further reduce output error.
% and the layer-wise quantizer optimization has been investigated.
% LLM.int8()
LLM.int8()~\cite{dettmers2022llmint8} uses a mixed-precision quantizer to isolate the outliers into 16 bits and quantize other values into 8 bits. 
% mixed-precision decomposition
% to maintain model performance while reducing memory footprint.
% SmoothQuant
SmoothQuant~\cite{xiao2023smoothquant} smooths the activation before quantization to eliminate the outliers in different input channels, thus reducing quantization loss.
% eliminates the outliers by smoothing the activation values in quantization.
% and migrates the quantization difficulty from activation to weights.

% \vspace{-5pt}
% \subsubsection{Layer-wise Quantizer Optimization}
2) \textbf{Layer-wise quantizer optimization} adopts a calibration dataset to 
% build a better quantizer that 
align the outputs of the quantized model and float one, as shown in  \figref{fig:development-stages-of-PTQ}.
% studies
% AdaRound
AdaRound~\cite{nagel2020adaround} learns a differentiable function that takes values between 0 and 1 to realize the adaptive rounding by minimizing layer-wise output error.
% BitSplit
BitSplit~\cite{wang2020bitsplit} constructs quantized values bit-by-bit using a squared output error objective on the residual error.
% BRECQ
BRECQ~\cite{li2020brecq} reconstructs the outputs of quantized layers and residual blocks to learn the best step size and rounding variable of a quantizer.
% introduces Fisher information for building the objective of output error and learns the best quantization step size and rounding variable of layers within a single residual block jointly.
% EasyQuant
EasyQuant~\cite{wu2020easyquantvit} jointly optimizes both scales of activation and weights for each Layer, targeting the loss of the cosine similarity between original and quantized convolutional outputs.
% PTQ4ViT
PTQ4ViT~\cite{yuan2022ptq4vit} proposes the twin uniform quantization for activation with asymmetric distributions in ViTs and introduces a Hessian-guided metric to determine the scaling factors of quantizers layer-by-layer.
% APQ-ViT
APQ-ViT~\cite{ding2022apqvit} works on maintaining the power-law character of the softmax activations in ViTs.
% and proposes a calibration scheme that perceives the overall quantization disturbance in a block-wise manner.
RepQ-ViT~\cite{li2023repqvit} studies a scale-reparametrized quantizer to minimize output error while maintaining efficient inference.
% RepQ-ViT~\cite{li2023repqvit} introduces the scale-reparametrized quantizer which decouples the quantization and inference processes, for employing a complex channel-wise and logarithmic quantization while maintaining efficient inference.
% Outlier Suppression+ (OS+) channel-wise
Outlier Suppression+ (OS+)~\cite{wei2023outlier} measures the output change after quantization to pursue effective quantizing factors.
% adopts the Mean Squared Error (MSE) to measure the output change after scaling and quantizing weight and activation to pursue effective quantization factors.
% AWQ
AWQ~\cite{lin2023awq} finds the proper scaling factors to preserve the outliers in activation, thus minimizing output error.

% Using the calibration dataset to guide quantization can better align the outputs, but optimizing only the quantizer parameters is difficult to further reduce output error, and then the layer-wise parameter adjustment is investigated.

% \vspace{-5pt}
% \subsubsection{Layer-wise Parameter Optimization}
3) \textbf{Layer-wise parameter optimization} also utilizes a calibration dataset to minimize output error directly.
It updates both layer parameters and quantizer parameters to reduce output error further, as shown in  \figref{fig:development-stages-of-PTQ}. 
% Typical studies include \cite{hubara2021adaquant,frantar2022gptq,yao2022zeroquant,frantar2022obc}.
% studies
% AdaQuant
AdaQuant~\cite{hubara2021adaquant} proposes layer-by-layer optimization to learn the optimal quantization step size and update layer weights to minimize the layer output error.
% AdaQuant~\cite{hubara2021adaquant} adds a float bias tensor for weights and proposes a layer-by-layer optimization method to learn the optimal quantization step size and update the bias tensor to minimize the layer output error.
% AdaQuant~\cite{hubara2021adaquant} proposes a layer-by-layer optimization method that minimizes the error between the quantized layer output and the full-precision layer output.
% It learns the optimal quantization step size and adds a continuous bias tensor before weight quantization to indirectly update weights.
% ZeroQuant
ZeroQuant~\cite{yao2022zeroquant} uses layer-by-layer knowledge distillation to fine-tune the layer parameters.
% to align the outputs of layers in the quantized model to that of the corresponding layers in the float model.
% FlexRound
FlexRound~\cite{lee2023flexround} learns a different scaling factor for each pre-trained weight to reconstruct each layer or block output.
% and realize the per-tensor quantization for low performance degradation.
% OBC and GPTQ
OBC~\cite{frantar2022obc} expands the Optimizing Brain Surgery (OBS)~\cite{hassibi1993optimalbrainsurgeon,frantar2021mfac}
% ,singh2020woodfisher,frantar2021mfac} 
algorithm to general quantization scenes. 
It iteratively quantizes a portion of the weights into discrete ones and updates the remaining float weights to eliminate output error during the quantization process.
This quantizing process is time-consuming and infeasible for large models.
The following GPTQ~\cite{frantar2022gptq} proposes a fast and robust implementation of OBC for large model quantization.

These methods have achieved great success in aligning the outputs.
% and greatly improving the model task performance. 
% By combining techniques such as block-wise and mixed-precision techniques, even 4-bit quantization can be achieved, greatly reducing the memory requirement and computational cost in model inference.
% even with full-parameter fine-tuning on the calibration dataset, 
However, PTQ is difficult to recover the task performance from ultra-low-precision quantization with a large output error~\cite{frantar2022gptq,yao2022zeroquant}.
% compared with QAT.
% Besides, in layer-wise parameter optimization, PTQ requires fine-tuning the full model parameters and quantizer parameters. The cost is gradually approaching QAT and cannot be ignored~\cite{frantar2022obc,li2020brecq,liu2023pdquant}.

\vspace{-5pt}
\subsection{Bias-Related Studies}
\vspace{-5pt}
% 目前只有BC和noise都可以看作是BC的一种，其他的还得多看看文献再写
% LoRA
% Quant-Noise: 将量化和剪枝当作是在原始权值上增加噪声映射,提出使用随机区域噪声映射的QAT方式来训练量化模型模型,类似于Dropout的方式,发现在ConvNet和Transformer上都可以实现更高的模型精度.
% HAWQ和kravchik2019low都把quantization当作是一种添加到权值或者激活上的bias，
% 基于偏置的思想研究量化的想法由来已久，比如
The idea of optimizing quantization with bias has a long history.
% HAWQ~\cite{dong2019hawq} proposes the mixed-precision quantization for different layers of models and it applies the bias with different variances to estimate the sensitivities of layers.
HAWQ~\cite{dong2019hawq} applies bias with different variances to weights to simulate quantization and estimate the sensitivities of layers.
\cite{kravchik2019low} adds a constrained bias tensor to the float weight tensor to implement quantization and measure the output error caused by the bias.
% Quant-Noise~\cite{stock2021quantnoise} takes the quantization as a noise function and proposes the block-wise method which quantizes and trains a different random block subset of weights during each forward.
AutoRound~\cite{nagel2020adaround}, AutoQuant~\cite{hubara2021adaquant}, and BRECQ~\cite{li2020brecq} add a continuous bias variable to weights for adaptive weight optimization while minimizing the output error. 
\cite{banner2019post} found that applying bias correction to align the means of weights before and after quantization can reduce accuracy degradation.
% although increasing the quantization loss.
NoisyQuant~\cite{liu2023noisyquant} adds a noise bias on activation before quantization to flatten the activation distribution for less quantization loss.
% pseudo quantization training
The Pseudo Quantization Training (PQT) methods~\cite{shin2023nipq,savarese2022trainablenoise} add learnable noise parameters to floating-point values to address the non-differentiable problems in QAT. 
% These methods 
% We argue that flattening distribution by adding noise cannot reduce quantization loss, but it has the potential to reduce output error.

% These investigations add bias or noise to the floating-point tensor before quantization, to simplify the quantization parameter optimization and reduce quantization loss.
All of these studies employ bias before quantization to minimize quantization loss or achieve differentiable quantization.
However, adding bias before quantization can complicate the following computation.
% and affect computing efficiency as bias is usually a floating-point number. 
% In addition, since the quantization process is discrete, the optimization of the bias in these studies is non-convex and difficult to solve.
In addition, the optimal bias in these studies is difficult to obtain since the optimization of bias added before quantization is non-convex.
% There is almost no work considering adding bias to the output tensor to directly reduce output error.

% 这些工作表明了bias在量化中的重要作用，但是鲜有人关注到bias可以直接用于降低输出误差
% 这些工作表明了bias具有降低output error的潜力，但是据我们所知，目前还没有使用bias来直接降低输出误差的工作。
% 目前还没有工作直接讨论bias补偿对输出误差的优化。
% 在本文中，我们显示地提出基于偏置补偿的量化输出误差优化，主要基于两个动机：
% 1、目前的PTQ方法即使使用全参数微调也难以实现更低位宽的量化，其发展陷入瓶颈。
% 2、偏置补偿具有进一步降低输出误差的潜力，但是还没有相关工作对其进行探索
% These studies show the potential for reducing output error in quantization.
% However, to the best of our knowledge, 
% there is no work arguing the application of bias compensation for reducing output error.

\begin{figure*}[!ht]
    \centering
    \includegraphics[width=1\textwidth]{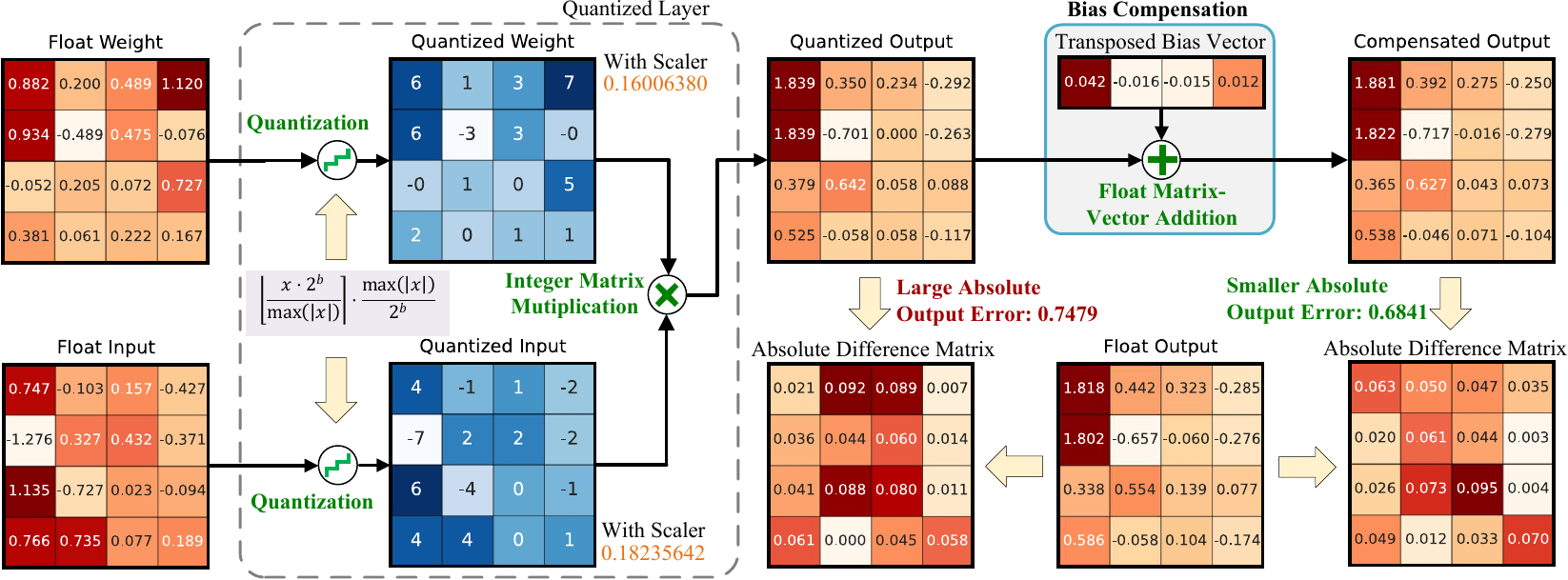}
    \vspace{-10pt}
    \caption{Illustration of bias compensation. 
    We use absolute error as output error for easy understanding.
    Applying bias compensation after quantization can significantly reduce the output error without increasing additional computational complexity.}
    \label{fig:illustration-of-bias-compensation}
\end{figure*}

\vspace{-5pt}
\subsection{Motivations}
We propose BC mainly driven by two motivations:
1) PTQ methods face challenges in achieving low-precision quantization due to significant output error.
2) Applying bias to output has the potential to reduce output error further, but to the best of our knowledge, relevant work has yet to explore it.
Therefore, we propose BC to address these issues. 
We prove that the minimization of output error through BC is convex, and its optimal solution can be easily obtained.
% Besides, our method is orthogonal to previous optimizations. 
% It can be easily combined with the existing quantizers and enables ultra-low precision quantization.

% 我们提出的偏置补偿具有以下优势：
% 1、搜索成本低，凸优化，可以有效地降低输出误差
% 2、效果好，与之前的量化优化正交，可以方便地与之前的工作集成

% \gc{Historys-detail}
% \cite{AutoQNN}
% \gc{Relate works-detail}

% \gc{Motivations-detail}

% 本章内容可以编写边看边总结，可以先写正文的内容

\vspace{-10pt}
\section{Methodology}
In this section, we first introduce the quantization and output error, which is the essential issue degrading model task performance.
Then we propose bias compensation from a novel perspective to reduce the output error.

\subsection{Quantization and Output Error}
% 在本段中，我们需要体现的是，量化导致的精度下降是由输出损失导致的
Let a neural network has $L$ layers, and the $i$-th layer with float weight $W_i\in \R^{m\times n}$ processes the float input $X_i\in \R^{b\times m}$, where $b$ is the min-batch size, $m$ is the feature size of each row in $X$, and $n$ is the output feature size of $i$-th layer.
The typical computation of $i$-th layer is as follows:
\begin{equation}
    X_{i+1}=X_i W_i.
\end{equation}
The matrix multiplication of $X_i W_i$ is the typical computational bottleneck in DNNs, and the quantization is proposed to reduce its computational load.

Let a quantizer $Q_{W_i}:\R^{m\times n}\rightarrow \Q^{m\times n}$ map any float value in $\R^{m\times n}$ into a quantized one in $\Q^{m\times n}$, \ie, quantized number set, and $Q_{X_i}$ for input $X_i$, the float matrix multiplication can be replaced with following implementation:
\begin{equation}\label{eq:quantized-matrix-multiplication}
\begin{split}
    X^q_{i+1}&=Q_{X_i}(X_i) Q_{W_i}(W_i)\\
    &~s.t.~Q_{X_i}(X_i)\in\Q^{b\times m}, Q_{W_i}(W_i)\in\Q^{m\times n}.
\end{split}
\end{equation}
Given that the space of $\Q$~is significantly smaller than that of $\R$, it becomes feasible to represent elements in $\Q$~using fewer bits.
Consequently, operations on these elements can be executed using low-bit integer operations, as shown in \figref{fig:illustration-of-bias-compensation}, leading to reduced computational load.
% a cheap quantized

However, the smaller space of $\Q$~implies inevitable numerical errors, 
and the error matrix can be defined as follows:
\begin{equation}\label{eq:quantization_errors}
    \begin{cases} 
        \N_{X_i} &= X_i-Q_{X_i}(X_i), \\
        \N_{W_i} &= W_i-Q_{W_i}(W_i).
    \end{cases}
\end{equation}
% The loss is usually defined as the Mean Square Error (MSE) between the float input and the quantized one.
% \begin{equation}\label{eq:l2_quantization_loss}
%     \begin{cases} 
%         l_{X_i} &= ||X_i-Q(X_i)||_2^2 \\
%         l_{W_i} &= ||W_i-Q(W_i)||_2^2
%     \end{cases}
% \end{equation}
Here, $\N_{X_i}$ and $\N_{W_i}$ are usually called noise matrixes since the numerical error of quantization can be regarded as applying a noise for each element in $X_i$ and $W_i$~\cite{dong2019hawq,nagel2020adaround,hubara2021adaquant}.
% 在许多PTQ量化算法中，噪声矩阵的L2范数通常被作为量化误差进行优化，以降低量化导致的模型任务性能下降。
In many previous studies~\cite{banner2019post,zhao2019improving,kravchik2019low}, the L2 norms of $\N_{X_i}$ and $\N_{W_i}$, \ie, $||\N_{X_i}||_2^2$ and $||\N_{W_i}||_2^2$, are optimized for improving the task performance of quantized models.

% Based on optimizing the above losses, the quantizer $Q$ can be corrected.
Recent studies show that aligning the outputs of quantized layers and their corresponding float layers is the pivot factor for high model task performance~\cite{yao2022zeroquant,frantar2022obc,frantar2022gptq}.
The difference between the outputs of $X_{i+1}$ and $X^q_{i+1}$ is defined as follows:
\begin{equation}\label{eq:output_errors}
    \N_{i+1} = X_{i+1}-X^q_{i+1}.
\end{equation}
Here $\N_{i+1}\in\R^{b\times n}$ is a difference matrix, and each element in it shows the error caused by the quantization processes $Q_{X_i}$ and $Q_{W_i}$  as shown in \figref{fig:illustration-of-bias-compensation}. 
Our target is to minimize the L2 norm of the difference matrix $\N_{i+1}$ as follows:
% Besides, what we are concerned about is the output difference between $X_{i+1}$ and $X^q_{i+1}$ instead of the above losses, and output error can be defined as follows.
\begin{equation}\label{eq:l2_output_loss}
\begin{split}
    l_{i+1} &= ||\N_{i+1}||_2^2,\\
    &=||\N_{X_i}\N_{W_i}-\N_{X_i}W_i -X_i\N_{W_i}||_2^2.
\end{split}    
\end{equation}
$l_{i+1}$ is the output error of $i$-th layer, which is only decided by noise matrixes $\N_{X_i}$ and $\N_{W_i}$.
To our best knowledge, almost all previous studies focus on finding the relatively optimal noise matrixes $\N_{X_i}$ and $\N_{W_i}$ that meet quantization constraints while minimizing the output error $l_{i+1}$:
\begin{equation}
\begin{split}\label{eq:l2_output_loss_optimization}
    &\underset{\N_{X_i},\N_{W_i}}{\arg\min}||\N_{i+1}||_2^2\\
    &~s.t.~\forall s \in (X_i-\N_{X_i}) \cup (W_{i}-\N_{W_i}), s \in \Q
\end{split}
\end{equation}
However, this optimization is non-convex since the quantization process is discrete.
It is difficult to solve the optimal noise matrixes for minimal output error.
% thus affecting model task performance.
% Generally, optimizing the quantization losses can reduce the output loss, and guide the quantizer design.  

\subsection{Bias Compensation}
% 做误差纠正的有两类，一类是在权值上做bias误差纠正，一类是在激活值上做误差纠正，但是据我们所知，目前还没有在输出上做误差纠正的工作。
Instead of optimizing the noise matrixes $\N_{X_i}$ and $\N_{W_i}$, we innovatively introduce the bias compensation to minimize output error.
% We introduce the bias compensation instead of optimizing the noise matrixes $\N_{X_i}$ and $\N_{W_i}$.
% Despite many studies having mentioned using bias in quantization as described in \secref{sec:relatedwork}, to our best knowledge, this is the first study to discuss bias compensation for optimizing output error in quantization. 
% 虽然之前有很多bias相关的工作，但是据我们所知，我们是第一个使用偏置补偿来优化输出误差的研究，
% Like the quantization errors, we first define the output errors as follows. 
% We define the output difference matrix as follows.
% \begin{equation}\label{eq:output_errors}
%     \N_{i+1} = X_{i+1}-X^q_{i+1}.
% \end{equation}
% Here $\N_{i+1}\in\R^{b\times n}$ is the difference matrix, and each element in it shows the error caused by the quantization processes $Q_{X_i}$ and $Q_{W_i}$. 
% Our target is to minimize the L2 norm of the difference matrix $\N_{i+1}$ as shown in \eqnref{eq:l2_output_loss}.
% The previous studies usually optimize the noise matrixes or design quantizers cautiously, but they are hard to get a feasible solution since these optimizations are non-convex.
% 23年：改到这里，明年再改
% Differing from the previous methods, 
Our idea is to find an extra bias vector to compensate for the quantized output matrix $X^q_{i+1}$ as follows:
\begin{equation}\label{eq:bias_compensation}
    \hat{X}^q_{i+1}(j)=X^q_{i+1}(j)+\B_{i+1}~~s.t.~j=1,2,\cdots,b.
\end{equation}
Here $\B_{i+1}\in\R^{n}$ is a bias vector, and $X^q_{i+1}(j)$ is the $j$-th row of matrix $X^q_{i+1}$.
% 这里我们需要着重强调一下这么做的动机：不改变量化过程和量化后的计算方式，计算和存储需求非常低，可以有效降低输出误差。
Attaching the bias vector $\B_{i+1}\in\R^{n}$ to $X^q_{i+1}$ follows the quantization process and quantized matrix multiplication as shown in \figref{fig:illustration-of-bias-compensation}.
It does not alter the quantization and computation procedures of quantized layers.
% thus will not change the quantization and computation processes of quantized layers.
Besides, the addition of bias vector and output matrix is an efficient operation that does not substantially amplify computational load while offering the potential for reduced output error.
With bias compensation for the quantized output, the output error can be reformulated as follows:
\begin{equation}\label{eq:bias_compensation_error}
\begin{split}
    l_{i+1}\!=\!||X_{i+1}-\hat{X}^q_{i+1}||_2^2 
            \!=\!\sum_{j=1}^b||\N_{i+1}(j)-\B_{i+1}||_2^2.
\end{split}    
\end{equation}
Here $\N_{i+1}(j)$ is the $j$-th row of matrix $\N_{i+1}$.
% 不同于以往的量化优化，我们的想法是将输出误差的优化当作是与之前量化过程正交的一个优化过程，将量化输出当作已知量，来优化偏置向量B。
% The quantization as an orthogonal process to bias compensation, 
% and we do not optimize $X^q_{i+1}$ in the bias compensation and see the $X_{i+1}^q$ as a constant matrix in \eqnref{eq:bias_compensation}.
Differing from the previous optimization target in \eqnref{eq:l2_output_loss_optimization}, our target is to find an optimal bias vector $\B_{i+1}$ for the quantized output of $i$-th layer, and minimizes the output error $l_{i+1}$:
\begin{equation}\label{eq:bias_optimization}
    \underset{\B_{i+1}}{\arg\min}\sum_{j=1}^b||\N_{i+1}(j)-\B_{i+1}||_2^2.
\end{equation}
% 值得注意的是，difference matrix N完全由量化决定，而BC是与量化过程正交的，因此，优化BC的过程是独立于量化优化的过程的，也就是说l最优的结果实际上应该是等价于优化difference matrix N的L2范数+优化N-B的L2范数这两个过程，只有这两个过程都达到最优，则l能达到最优。而就目前而言，前者的优化是非凸的，而后者的优化是凸优化。
Notably, any value of $\B_{i+1}$ will not affect $\N_{i+1}$, thus the optimization in~\eqnref{eq:bias_optimization}~is irrelated to the optimization in \eqnref{eq:l2_output_loss_optimization}.
It implies that bias compensation can be easily combined with various quantizers for lower output error.
% 当考虑quantizer设计时，最优的输出误差应该是eq1和eq2同时达到最优，然而eq1是非凸的，在现有条件下无法找到最优解。然而我们接下来会证明eq2是平凡的，可以求得最优解
% Besides, 

\section{Optimization}\label{sec:implementation}
% 实现章节：
% 3、凸优化证明
% 4、最优偏置向量求解
% 5、偏置补偿的应用位置和算法
% 我们会证明，无论给定任意的quantizer，应用BC都能保证更低的输出误差，证明||N-B||<=||N||
In this section, we present a proof demonstrating that the optimal bias vector of BC can be solved and it always guarantees a lower output error.
To do this, we first prove that the minimization of output error with respect to the bias vector is convex, and then solve the optimal bias vector based on a given quantizer and calibration dataset.
Finally, we present the inference of a Transformer layer with BC.
% in a typical Transformer architecture.

\subsection{Convex Optimization Proof}
% 添加一节来证明添加bias一定会降低输出误差而不会增加。理论证明一下。其实就是把这个最优的B带入进去看li+1是不是会达到更小的值。
% 计算该函数的Hessian矩阵来证明该函数是凸函数，基于该函数的优化是凸优化
Let $\mathbf{H}$ be the Hessian matrix of $l_{i+1}$ with respect to $\B_{i+1}$ in \eqnref{eq:bias_compensation_error}, 
the elements in $\mathbf{H}$ are computed as follows.
\begin{equation}
   \mathbf{H}(p,q)=\!\frac{\partial^2 l_{i+1}}{\partial \B_{i+1}(p)\partial \B_{i+1}(q)}\!=\!\begin{cases}
        b>0, p = q\\
        0,p\neq q
    \end{cases}.
\end{equation}
Here $\mathbf{H}(p,q)$ refers to the element in the $p$-th row and $q$-th column of $\mathbf{H}$.
Clearly, each element in $\mathbf{H}$ is greater than or equal to 0,
which indicates that $\mathbf{H}$ is positive-semidefinite and the optimization in \eqnref{eq:bias_optimization} is convex.

\begin{figure}[!t]
    \centering
    \includegraphics[width=1\columnwidth]{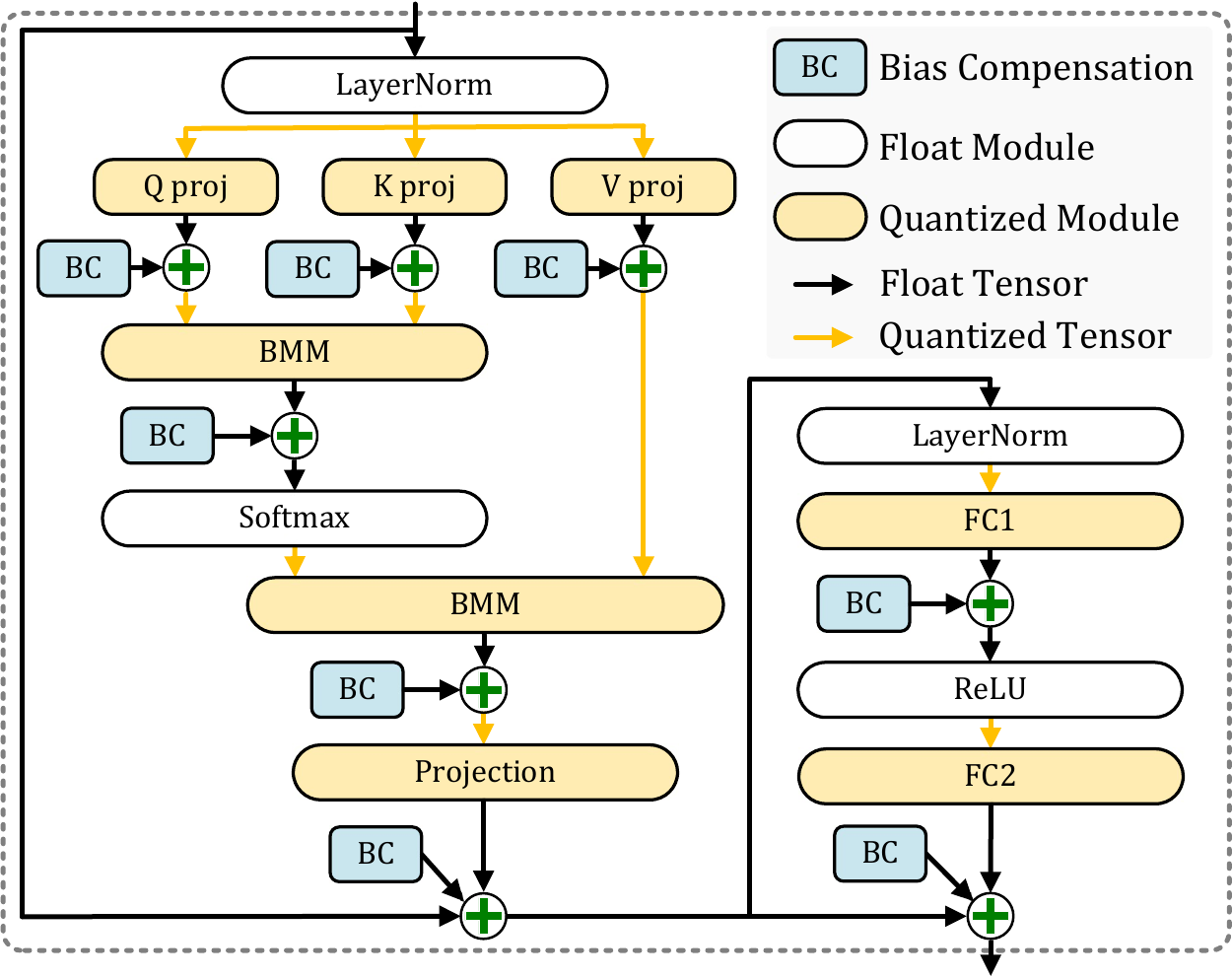}
    \vspace{-15pt}
    \caption{Bias compensation positions in Transformer.}
    \label{fig:bc-positions-in-transformer}
\end{figure}
\subsection{Obtain Optimal Bias Vector}
For a convex optimization, we can solve for its optimal analytical solution directly.
Let the optimal bias vector be $\B_{i+1}^*$ and the optimal solution is obtained as follows:
\begin{equation} 
    \B_{i+1}^*\!\!=\!\frac{1}{b}\sum_{j=1}^b \N_{i+1}(j)\!=\!\frac{1}{b}\sum_{j=1}^b (X_{i+1}(j)-X^q_{i+1}(j)).
\end{equation}
Here $b$ refers to the number of samples of a calibration dataset, and $X_{i+1}$ and $X^q_{i+1}$ are certain for a given quantizer and input.
The optimal bias vector $\B_{i+1}$ for $i$-th quantized layer can be obtained with two feedforward inferences (computing $X_{i+1}$ and $X^q_{i+1}$) without fine-tuning.
% 然后带入求解的B回去证明，最优的||N+B||<=||N||

% We can prove that employing BC for any quantizer and calibration dataset can gurantee the lower output error as follows.
Employing the optimal bias vector $\B_{i+1}^*$, the lowest output error can be obtained as follows.
\begin{equation}\label{eq:gurantee_proof}
    \begin{split}
        l_{i+1} & = \sum_{k=1}^b||\N_{i+1}(k)-\B_{i+1}^*||_2^2 \\
        &=\sum_{k=1}^b||\N_{i+1}(k)-\frac{1}{b}\sum_{j=1}^b \N_{i+1}(j)||_2^2 \\
        &=||\N_{i+1}||_2^2-b||\frac{1}{b}\sum_{j=1}^b \N_{i+1}(j)||_2^2 \\
        &\leq ||\N_{i+1}||_2^2.
    \end{split}
\end{equation}

It demonstrates that \emph{BC can always guarantee the lower output error} with a given quantizer and calibration dataset.

% \usepackage{multirow}
% \renewcommand{\arraystretch}{1.1}
\begin{table*}[!ht]
\centering
\caption{Top-1 accuracy results of ViTs, Deits, and Swins. 
The default input resolution is 224 × 224 ($^*$ means 384 × 384).
% BC greatly improves the accuracy of ViTs using PTQ4ViT quantization across all bits and datasets and enables PTQ4ViT to achieve competitive accuracy in ultra-low-precision quantization without the need for any updates of the quantizer parameters and weights, or fine-tuning. 
% BC greatly improves the accuracy of ViTs using PTQ4ViT quantization across all bits and datasets, and
BC enables PTQ4ViT to achieve competitive accuracy in ultra-low-precision quantization without quantizer parameter or layer weight
optimization. 
}
\label{tab:combined-results}
\setlength\tabcolsep{4.4pt} % default value: 6pt
\renewcommand{\arraystretch}{1.0}
\begin{tabular}{l|c|ccc|ccc|cccccc}
\hline
Model & W/A 
& \multicolumn{1}{l}{ViT-S}
%& \multicolumn{1}{l}{ViT-S$^*$} 
& \multicolumn{1}{l}{ViT-B} 
& \multicolumn{1}{l|}{ViT-B$^*$}
& \multicolumn{1}{l}{Deit-S}
& \multicolumn{1}{l}{Deit-B} 
& \multicolumn{1}{l|}{Deit-B$^*$}
& \multicolumn{1}{l}{Swin-T}
& \multicolumn{1}{l}{Swin-S} 
& \multicolumn{1}{l}{Swin-B}
& \multicolumn{1}{l}{Swin-B$^*$} \\\hline
\# Parameter(M) 
& - & 22.1&86.6 &86.9
& 22.4& 86.6&86.9
& 28.3	&49.9	&88.1	&90.8  \\\hline
FP32 & 32/32& 81.39 & 84.54 & 86.00  
& 79.85 &81.80  & 83.11 
& 81.39 &83.23 &85.27& 86.44 \\\hline
Base PTQ & 8/8 &  80.46 & 83.89 & 85.35 
& 77.65 &80.94 & 82.33 
& 80.96 &82.75 & 84.79 & 86.16 \\
EasyQuant & 8/8 & 80.75 & 83.89 & 85.53 
& 78.98& 81.19 &82.10 
& 80.95 &83.00 &85.10 &86.39  \\
PTQ4ViT & 8/8 & 81.00 & 84.25 & 85.82
& 79.47 &81.48& 82.97
& 81.24 &83.10 &85.14& 86.39\\
APQ-ViT & 8/8 & \textbf{81.25}& 84.26 & -
& \textbf{79.78}& 81.72 & -
& - & 83.16& 85.16& 86.40\\
NoisyQuant & 8/8 & 81.15& 84.22 & 85.86 
& 79.51& 81.45 &82.49
& 81.25& 83.13& 85.20& \textbf{86.44}\\
\textbf{PTQ4ViT+BC}& 8/8 & 81.15 & \textbf{84.33} & \textbf{85.96} 
& 79.58 & \textbf{81.76} & \textbf{83.12} 
& \textbf{81.34}&	\textbf{83.18}&\textbf{85.21}&\textbf{86.44}
 \\\hline
Base PTQ & 6/6 & 70.24 & 75.66 & 46.88 
& 72.26& 78.78 &68.44 
& 78.45& 81.74 & 83.35 & 85.22 \\
EasyQuant & 6/6 & 75.13 &81.42 & 82.02 
& 75.27 &79.47 &81.26
& 79.51 &82.45 &84.30 &85.89\\
PTQ4ViT & 6/6 & 78.63 &  81.65 & 83.34
& 76.28 &80.25& 81.55 
& 80.47 &82.38 &84.01& 85.38 \\
APQ-ViT & 6/6 &
79.10&
82.21 & - & 
77.76& 80.42& - & - &
82.67 &84.18 &85.60\\
NoisyQuant & 6/6 & 78.65 & 82.32 & 83.22 
& 77.43& 80.70& 81.65 
& 80.51& \textbf{82.86}& \textbf{84.68} &\textbf{86.03} \\
RepQ-ViT & 6/6 &
\textbf{80.43}&
\textbf{83.62} & - &
\textbf{78.90}&
81.27& - & - &
82.79&
84.57& - & \\
\textbf{PTQ4ViT+BC}& 6/6 & 79.22 & 83.00& \textbf{85.00} 
& 78.60 & \textbf{81.29} & \textbf{82.44} 
& \textbf{80.56}&	82.46&84.28&	85.60
 \\\hline
PTQ4ViT & 4/4 & 34.02 & 35.30 & 31.40
& 24.05 & 60.72 & 75.93
& 74.46&	76.46	&74.49&77.26
\\
APQ-ViT & 4/4 &
47.95&
41.41 & - &
43.55&
67.48& - & - &
77.15&
76.48& \textbf{80.84}&  \\
RepQ-ViT & 4/4 &
\textbf{65.05}&
\textbf{68.48} & - &
\textbf{69.03}&
75.61& - & - &
\textbf{79.45}&
\textbf{78.32}& - &  \\
\textbf{PTQ4ViT+BC} & 4/4 & 54.74  & 67.27 & \textbf{68.29}  
& 67.47	&\textbf{75.78}	&\textbf{80.15}	& \textbf{74.74}	&76.79	&77.78	&79.97
\\\hline
\end{tabular}
\end{table*}
\subsection{Inference with Bias Compensation}
In the implementation of model inference using BC, the bias vector is attached after each quantized layer in order to reduce output error.
In a typical Transformer architecture, the fully connected layers and batch matrix multiplication (BMM) layers are quantized to minimize computational costs.
The BC modules are then embedded after each quantized module, as illustrated in \figref{fig:bc-positions-in-transformer}.
Compared to the basic quantized model inference process, the quantized layers with BC modules only require an additional 
% memory allocation of bias vector and a 
matrix-vector addition.
These modules do not impact the computational efficiency of the quantized layers since the bias vectors are added to the outputs after the expensive operations.
BC has minimal impact on the computational and spatial complexity of quantized models, while ensuring lower output error with the given quantizer and calibration dataset.

\section{Experiments}

\subsection{Experiments on ViTs}
% ViTs demonstrate high accuracy with large parameters and datasets. We verify the proposed BC on ViTs in this subsection.
% ViTs have demonstrated that pre-training Transformer architectures with a large number of parameters on a massive visual dataset can achieve high accuracy. 
% However, 
Achieving low-precision ViTs remains a challenge.
In this experiment, we aim to demonstrate that BC can significantly enhance the accuracy of ViTs under low-bit quantization.
\subsubsection{Baselines}
We take the PTQ4ViT~\cite{yuan2022ptq4vit} as the base quantizer for BC and the \sArt~quantizers including Base PTQ in~\cite{yuan2022ptq4vit}, EasyQuant~\cite{wu2020easyquantvit}, APQ-ViT~\cite{ding2022apqvit}, NoisyQuant~\cite{liu2023noisyquant}, and RepQ-ViT~\cite{li2023repqvit} as baselines for comparison.
% 添加一点点说明。
% The open implementations of PTQ4ViT\footnote{github links for and EasyQuant} are adopted in this experiment for reproducibility. 
The experimental results of the baselines are cited from their original paper, and 
the results of PTQ4ViT are obtained based on the open implementation\footnote{https://github.com/hahnyuan/PTQ4ViT}.

\subsubsection{Models \& Datasets}
We evaluate BC on ImageNet-1K~\cite{deng2009imagenet} dataset with ViT~\cite{dosovitskiy2020vits}, DeiT~\cite{touvron2021deit}, and  Swin~\cite{liu2021swin} models.
All the pre-trained full-precision models and the default data augmentation strategies are cited from Timm~\cite{rw2019timm}.
% The ImageNet-1K~\cite{deng2009imagenet} dataset is used and top-1 accuracy on the ImageNet-1K validation dataset is metric. 
We randomly select 32 images from the ImageNet-1K training dataset as calibration images in the model quantization.
% as did in~\cite{yuan2022ptq4vit}. 
% We randomly selected 128, 64, and 32 images for calibration, and then selected the best results.
% The default data augmentation in~\cite{rw2019timm} for ImageNet-1K is adopted.

% \usepackage{multirow}
% \renewcommand{\arraystretch}{1.1}
\begin{table*}[!ht]
\centering
% \usepackage{multirow}
\caption{Perplexity results of LLMs. 
BC substantially reduces the perplexity of LLMs with GPTQ on various bits and datasets.
}
\label{tab:ppl-llms}
\setlength\tabcolsep{4.9pt} % default value: 6pt

\renewcommand{\arraystretch}{1.0}
\begin{tabular}{l|c|ccc|ccc|ccc}\hline
\multicolumn{1}{c|}{\multirow{2}{*}{Method}} & \multirow{2}{*}{Bits} & \multicolumn{3}{c|}{OPT-125M} & \multicolumn{3}{c|}{OPT-350M} & \multicolumn{3}{c}{BLOOM-560M} \\\cline{3-11}
\multicolumn{1}{c|}{} &  & WikiText2 & PTB & C4 & WikiText2 & PTB & C4 & WikiText2 & PTB & C4 \\\hline
FP16 & 16 & 27.66 & 32.55 & 24.61 & 22.00 & 26.08 & 20.71 & 22.42 & 41.26 & 24.38 \\\hline
RTN & 4 & 37.28 & 45.11 & 31.64 & 25.94 & 31.12 & 23.94 & 25.89 & 48.57 & 27.42 \\
GPTQ & 4 & 31.22 & 36.93 & 26.94 & 24.20 & 28.89 & 22.60 & 23.98 & 44.53 & 25.60 \\
\textbf{GPTQ+BC} & 4 & \textbf{29.90} & \textbf{36.11} & \textbf{26.03} & \textbf{22.85} & \textbf{27.23} & \textbf{21.34} & \textbf{23.77} & \textbf{44.44} & \textbf{25.46} \\\hline
RTN & 3 & 1276.92 & 1209.34 & 731.60 & 64.56 & 81.85 & 50.14 & 56.98 & 117.15 & 58.96 \\
GPTQ & 3 & 54.68 & 65.74 & 38.27 & 33.75 & 38.66 & 28.53 & 32.45 & 62.66 & 32.25 \\
\textbf{GPTQ+BC} & 3 & \textbf{45.19} & \textbf{52.15} & \textbf{33.78} & \textbf{27.78} & \textbf{33.79} & \textbf{24.79} & \textbf{31.49} & \textbf{60.36} & \textbf{31.35} \\\hline
RTN & 2g64 & 7042.44 & 5057.60 & 3869.38 & 4354.61 & 3560.32 & 2346.04 & 502.39 & 627.17 & 326.20 \\
GPTQ & 2g64 & 192.96 & 200.53 & 114.07 & 519.97 & 506.01 & 231.55 & 74.06 & 182.64 & 59.23 \\
\textbf{GPTQ+BC} & 2g64 & \textbf{108.98} & \textbf{137.12} & \textbf{70.11} & \textbf{88.09} & \textbf{95.57} & \textbf{56.17} & \textbf{69.12} & \textbf{178.73} & \textbf{56.43}\\\hline
\end{tabular}
\end{table*}
\subsubsection{Exprimental Setting}
% The same as~\cite{liu2021posttrainingquantization,yuan2022ptq4vit}, 
We quantize all the weights and inputs for all the fully-connect layers and the matrix multiplications in self-attention modules in ViTs. 
Then, we attach BC modules following these quantized layers, as shown in \figref{fig:bc-positions-in-transformer}. 
% The softmax and normalization layers in ViTs are not quantized as did in \cite{liu2021posttrainingquantization,yuan2022ptq4vit}.
% as shown in \figref{fig:bc-positions-in-transformer}.
% 说一下BC的具体配置，比如输入时B,S,T，分别对应Batchsize， Sequence， 和Token维度，bias vector是和哪个维度对齐的？
The layer outputs in ViTs are typically three-dimensional tensors, encompassing batch, sequence, and token dimensions.
The size of the bias vector of BC module is identified by the product of the token dimension size and the sequence dimension size. 
During model inference, the bias vector is added to each batch slice of the output tensor.

\subsubsection{Results \& Analysis}
The accuracy results are shown in \tabref{tab:combined-results}.
BC consistently outperforms the base quantizer, \ie, PTQ4ViT~\cite{yuan2022ptq4vit}, across all bit quantization and models, and significantly improves the model accuracy.
The average accuracy improvements are $\textbf{0.12\%}$, $\textbf{0.85\%}$, and $\textbf{15.89\%}$ for W8A8, W6A6, and W4A4 quantizing settings, respectively, across all models.
It is attributed to the fact that PTQ4ViT often results in a large output error when using low-precision quantization.
BC minimizes the error and aligns the float outputs and quantized outputs, thus improving model accuracy.
% 越低精度的量化会导致越大的输出误差，而BC可以最大程度地缓解输出误差，针对大的输出误差，其表现更突出。
% BC在W4A4的超低位宽量化设置下的表现令人惊叹
Remarkably, BC enables low-precision quantization of PTQ4ViT without quantizer parameter or layer weight optimization.
% For instance, BC improve the accuracy of ViT-B$^*$ with 4-bit PTQ4ViT quantization by up to $\textbf{34.80\%}$, 
For instance, ViT-B* with 4-bit PTQ4ViT provides an accuracy of $31.40\%$. 
BC improves it by $\textbf{36.89\%}$, achieving an accuracy of $68.29\%$.
% Besides, the results of BC under the ultra-low bit quantization setting of W4A4 are surprising and inspiring.
% Besides, the lower precision quantization can result in larger output error, and applying BC in ultra-low precision quantization can significantly alleviate the error and improve accuracy.
% Especially, the proposed BC can largely improve the accuracy of models with ultra-low precision quantization.
% The accuracy improvement is up to $\textbf{33.49\%}$ when quantizing ViT-B$^*$ into 4-bit weight and activation.
% The results demonstrate that BC can enable ultra-low-precision quantization.

\begin{figure}[!t]
    \centering\includegraphics[width=1\columnwidth]{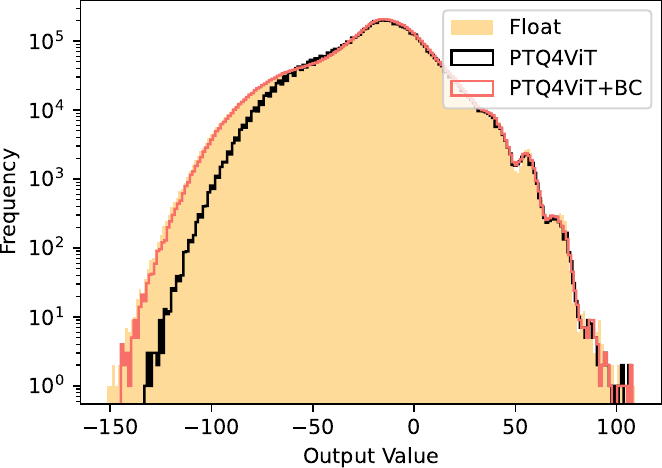}
    \vspace{-15pt}
    \caption{
    The attention output distribution of the first layer of ViT-S Transformer using 4-bit quantization.
    The output of PTQ4ViT significantly deviates from the float output.
    BC compensates for the output of PTQ4ViT and aligns it with the float one.
    }
    \label{fig:hist-comparison-of-bc-and-ptq4vit}
    \vspace{-10pt}
\end{figure}
BC outperforms \sArt~baselines and achieves competitive accuracy results.
% Base PTQ
% EasyQuant
% APQ-ViT
Compared to Base PTQ~\cite{yuan2022ptq4vit}, EasyQuant~\cite{wu2020easyquantvit}, and APQ-ViT~\cite{ding2022apqvit}, BC consistently achieves higher accuracy across various bits and models. 
It can be attributed to the high base accuracy of PTQ4ViT, coupled with BC to reduce layer-wise output errors further.
% NoisyQuant
When compared to NoisyQuant~\cite{liu2023noisyquant}, BC outperforms it in terms of accuracy across all models except for some cases of Swin models, where the results are slightly lower by a maximum margin of 0.43\%.
% RepQ-ViT
BC demonstrates comparable or even higher accuracy than RepQ-ViT~\cite{li2023repqvit} on certain models and bits, showcasing competitive performance.
However, it yields lower results in certain cases.
This discrepancy primarily arises from the significant contrast in accuracy between the quantizers, PTQ4ViT and RepQ-ViT. 
The optimization upper bound of BC remains constrained by the base quantizer, \ie, PTQ4ViT, as illustrated in \eqnref{eq:gurantee_proof}, and the large quantization error of the base quantizer can impact its optimization effectiveness and final achievable accuracy.
% Despite BC can significantly decrease the layer-wise output error, its optimization upper bound remains constrained by the base quantizer, \ie, PTQ4ViT, as illustrated in \eqnref{eq:gurantee_proof}.
% The notable quantization error of the base quantizer impacts the optimization effectiveness of BC.
Despite the limitations, BC greatly improves the task performance of the base quantizer and attains high accuracy. 
Remarkably, BC achieves this without requiring any updates of the quantizer parameters and layer weights, or the need for fine-tuning.
This robustly showcases the effectiveness of BC in ViT quantization.

% It improves the model accuracy from $34.80\%$, which is an infeasible accuracy in impractical, to $68.29\%$, which can be deployed in practice. 
% The results demonstrate that BC can enable ultra-low-precision quantization.

% 如果有比原来的结果还差的,这里就需要解释一下为什么会比原来的差
To explicitly demonstrate the effectiveness of BC in aligning outputs and reducing output error, we present the attention output distribution of the first layer of the ViT-S Transformer with 4-bit quantization on the calibration dataset, as shown in \figref{fig:hist-comparison-of-bc-and-ptq4vit}. 
It can be seen that the output of PTQ4ViT diverges significantly from the float output, resulting in a substantial output error. 
Remarkably, BC 
% introduces a bias vector to 
can compensate for the deviations and align the PTQ4ViT's output with float output, thereby considerably reducing the output error.

\subsection{Experiments on LLMs}
% 分别使用Bert和GPT2作为代表模型进行量化测试吧
% method baseline就选择GPTQ就可以了
% LLMs have achieved impressive performance in recent years, but their deployments are extremely difficult due to the massive parameters and operations.
% Therefore, verifying the effectiveness of quantization on LLMs is crucial for promoting their practical application.
% In this subsection, 
We conduct experiments to verify the effectiveness of BC in LLM quantization.

\subsubsection{Baselines}
% The \sArt~quantizers for LLM quantization are selected as baselines, and they are GPTQ~\cite{frantar2022gptq}, AWQ~\cite{lin2023awq}, SmoothQuant~\cite{xiao2023smoothquant}, LLM.int8~\cite{dettmers2022llmint8}, and ZeroQuant~\cite{yao2022zeroquant}.
We take the naive Rounding-To-Nearest (NTR) (implemented in~\cite{frantar2022gptq}) and~GPTQ~\cite{frantar2022gptq} as baselines in this experiment.
% We cite the experimental results provided in their papers and
We use the open implementation\footnote{https://github.com/IST-DASLab/gptq} of GPTQ and pre-trained quantized LLM models in HuggingFace to obtain reproducible results.
% and ensure they are consistent with the results presented in ~\cite{frantar2022gptq}.

\subsubsection{Models \& Datasets}
% Since LLMs with a small number of parameters are more sensitive to low-precision quantization~\cite{frantar2022gptq, frantar2022obc}, i.e., the more significant task performance degradation, this experiment adopts these LLMs to highlight the effectiveness of BC.
% Specifically, 
The OPT-125M~\cite{zhang2022opt}, OPT-350M, and BLOOM-560M~\cite{workshop2022bloom} are adopted in this experiment and their open implementations in HuggingFace are utilized for reproducibility.
Similar to GPTQ, we randomly select 128 random 2048 token segments from the C4 dataset~\cite{raffel2020exploring} as calibration data to obtain the best bias vectors for compensation.
% The evaluation datasets include Wikitext2~\cite{wikitex-2}, PTB~\cite{ptb}, C4~\cite{Exploring the limits of transfer learning with a unified text-to-text transformer}, GLUE~\cite{GLUE}, and LAMBADA~\cite{The LAMBADA dataset: Word prediction requiring a broad discourse context}.
% For BERT, we use the GLUE task to evaluate the performance of the baselines.
% For other LLMs, we use perplexity as the evaluation metric to verify model task performance.
We profile the quantized models on perplexity evaluation on WikiText2~\cite{merity2016wikitext}, PTB~\cite{marcus1994penntreebank}, and C4~\cite{raffel2020exploring} since perplexity can stably reflect the LLM’s performance~\cite{dettmers2023case}.

\subsubsection{Experimental Setting}
Only weight quantization is performed and the activation maintains FP16 as did in~\cite{frantar2022gptq}.
We apply BC for all the quantized modules in LLMs.
The size of the bias vector is set to the product of the token dimension size and the sequence dimension size of the module's output.

\begin{figure}[!t]
    \centering\includegraphics[width=1\columnwidth]{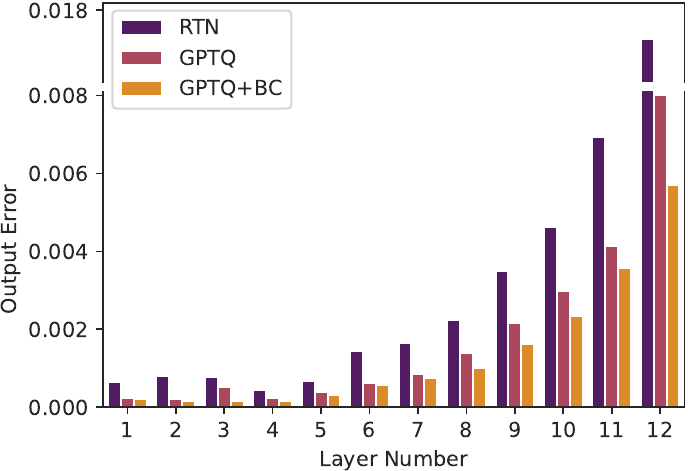}
    \vspace{-15pt}
    \caption{Layer-wise output errors of OPT-125M with different quantization methods on calibration dataset.
    BC greatly decreases the output errors across all layers.
    }
    \label{fig:output-error-comparison}
    \vspace{-10pt}
\end{figure}
\subsubsection{Results \& Analysis}
Perplexity results are shown in \tabref{tab:ppl-llms}.
The results of LLMs adopting BC consistently outperform that of baselines across all conditions.
% Besides, BC achieves significantly better perplexity results in the lower quantizing bits.
The average perplexity improvements across models and datasets are $\textbf{3.99}$ for 4-bit quantization, $\textbf{364.85}$ for 3-bit quantizing setting, and $\textbf{2912.83}$ for 2-bit (with a group size of 64) quantization, respectively.
% 更低位宽的提升效果越好，因为更低位宽量化的输出误差更大，BC可以显著地降低输出误差，从而回复模型的任务性能。
% 此外，由于较低精度的量化会导致较大的输出误差，并且BC擅长于此，因此偏置补偿在较低的量化比特中实现了明显更好的困惑结果。
% 更低量化位宽下的性能提升更明显，这表明应用BC有利于促进更低位宽的量化
These improvements can be attributed to the fact that BC can significantly reduce quantization output error and recover the perplexity of LLMs under low-precision quantization.
Besides, BC achieves significantly better perplexity results in the lower quantizing bits.
% since the lower precision quantization can result in larger output error.
It indicates that BC is beneficial for promoting ultra-low precision quantization in LLMs.
Especially, by applying BC, the perplexity of OPT-125M with 3-bit GPTQ has been reduced by $\textbf{13.59}$ on the PTB dataset, and that of OPT-350M with 4-bit GPTQ on the Wikitext2 achieves $\textbf{22.85}$, which is approaching the result of the float model ($22.00$).
BC significantly improves the results of LLMs in 2-bit quantization, with an order of magnitude improvement in many results. 
Compared to GPTQ, its average improvement reaches $\textbf{135.63}$.
These results demonstrate the significant potential of BC in suppressing output error and enhancing the task performance of quantized LLMs.

\begin{figure}[!t]
    \centering
    
    \includegraphics[width=1\columnwidth]{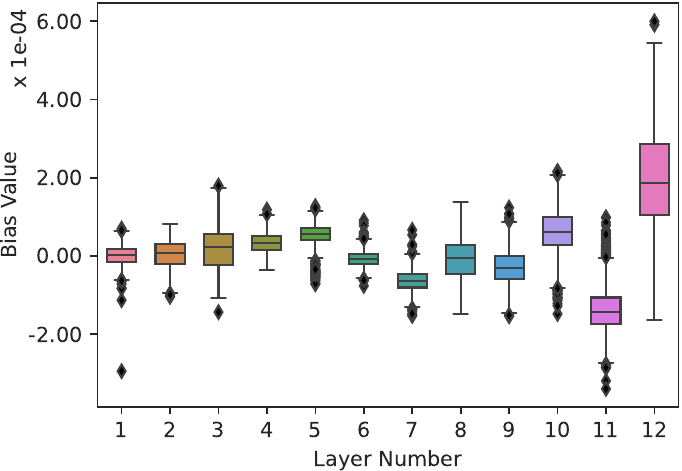}
    \vspace{-15pt}
    \caption{Distribution of bias values in different layers of OPT-125M using 3-bit GPTQ quantization with BC.
    Bias vectors for deeper layers have larger absolute means and variances since deeper layers exhibit greater output errors.
    }
    \label{fig:gptq-bias-distribution}
    \vspace{-10pt}
\end{figure}
% 把逐层的输出误差及其添加了BC的输出误差打印出来，证明输出误差降低，最好把RTN的结果也打印出来看看，作为对比。
% The performance improvement of quantized LLMs comes from the reduction of layer-wise output errors. 

To explicitly demonstrate the reduction of layer-wise output error, 
we compare the output errors of 12 Transformer layers of the OPT-125M model using 3-bit RTN, GPTQ, and GPTQ+BC, respectively, on the calibration dataset. 
As shown in \figref{fig:output-error-comparison}, 
BC greatly decreases the output errors across all layers in comparison to RTN and GPTQ.
% compared with RTN and GPTQ, BC can largely reduce the output errors across all layers.
It is consistent with the statement in \secref{sec:implementation}.
Besides, we observe that the deeper layers exhibit greater output errors, which is consistent with the previous observation~\cite{kravchik2019low}.
% ~\cite{Low-bit Quantization of Neural Networks for Efficient Inference}.
BC effectively compensates for these large output errors, thus improving the task performance of LLMs.

% 目的：展示bias的分布，说明给予GPTQ量化后的输出与全精度的输出还是存在显著偏差的，并且不同层的输出误差并不一样，因此对不同层应该应用不同的bias进行补偿。
% 根据之前的研究，量化对越深层的输出影响越大，即越深层可能导致与对应浮点层的输出误差越大，对此，我可以发现bias越深层的绝对均值越大，方差越大，即对深层的输出补偿更大的偏差，这与之前的研究发现一致\cite{}，
We present the distribution of bias vectors of BC modules for the multi-head attention layers in the OPT-125M model with 3-bit quantization, as shown in \figref{fig:gptq-bias-distribution}.
Noticeably, different bias vectors are obtained for different layers for low output errors, and the bias vectors for deeper layers have larger absolute means and variances, reflecting that the deeper layers have larger output errors, as described above.

% \gc{Our method setting}

% \gc{Metrics setting and experiments' target}

% \gc{Comparison platform: data sets, models, conditions}

% \gc{Comparison methods list: include their analysis}

% \gc{Results and analysis}

\section{Conclusion}
In this paper, we introduce a novel method called BC aimed at minimizing the output error induced by quantization in neural networks.
We theoretically prove that BC optimization is convex and that the optimal solution for minimal output error can be solved without the need for fine-tuning.
Additionally, we theoretically demonstrate that applying BC can always guarantee a lower output error with a given quantizer and calibration dataset.
Through extensive experiments conducted on ViTs and LLMs, the effectiveness of BC is highlighted in significantly enhancing the task performance of quantized models, particularly for ultra-low-precision quantizing settings of 4 bits, 3 bits, and even 2 bits.

% \gc{Work conclusion}

% \gc{Main idea}

% \gc{Defects and future works}

\bibliographystyle{icml2024}
\bibliography{references}

%%%%%%%%%%%%%%%%%%%%%%%%%%%%%%%%%%%%%%%%%%%%%%%%%%%%%%%%%%%%%%%%%%%%%%%%%%%%%%%
%%%%%%%%%%%%%%%%%%%%%%%%%%%%%%%%%%%%%%%%%%%%%%%%%%%%%%%%%%%%%%%%%%%%%%%%%%%%%%%
% APPENDIX
%%%%%%%%%%%%%%%%%%%%%%%%%%%%%%%%%%%%%%%%%%%%%%%%%%%%%%%%%%%%%%%%%%%%%%%%%%%%%%%
%%%%%%%%%%%%%%%%%%%%%%%%%%%%%%%%%%%%%%%%%%%%%%%%%%%%%%%%%%%%%%%%%%%%%%%%%%%%%%%
% \newpage
% \appendix
% \onecolumn

% You can have as much text here as you want. The main body must be at most $8$ pages long.
% For the final version, one more page can be added.
% If you want, you can use an appendix like this one.  

% The $\mathtt{\backslash onecolumn}$ command above can be kept in place if you prefer a one-column appendix, or can be removed if you prefer a two-column appendix.  Apart from this possible change, the style (font size, spacing, margins, page numbering, etc.) should be kept the same as the main body.
% %%%%%%%%%%%%%%%%%%%%%%%%%%%%%%%%%%%%%%%%%%%%%%%%%%%%%%%%%%%%%%%%%%%%%%%%%%%%%%%
% %%%%%%%%%%%%%%%%%%%%%%%%%%%%%%%%%%%%%%%%%%%%%%%%%%%%%%%%%%%%%%%%%%%%%%%%%%%%%%%

\end{document}